\documentclass{article}
\usepackage[final]{corl_2022} 
\usepackage{times} 
\usepackage{brian}
\graphicspath{{images/}}
\usepackage[export]{adjustbox}
\usepackage{array}
\usepackage{wrapfig}
\usepackage{xcolor}
\newcommand{\changed}[1]{{\color{black} #1}}

\newcolumntype{R}[2]{%
    >{\adjustbox{angle=#1,lap=\width-(#2)}\bgroup}%
    l%
    <{\egroup}%
}
\newcommand*\rot[2]{\multicolumn{1}{R{#1}{#2}}}

\title{
  Data-Efficient Model Learning for Control with Jacobian-Regularized Dynamic-Mode Decomposition}

\author{
Brian E. Jackson \\
Robotics Institute \\
Carnegie Mellon University \\
\texttt{brianjackson@cmu.edu} \\
\and
Jeong Hun Lee \\
Robotics Institute \\
Carnegie Mellon University \\
\texttt{jeonghunlee@cmu.edu} \\
\and
Kevin Tracy \\
Robotics Institute \\
Carnegie Mellon University \\
\texttt{ktracy@cmu.edu} \\
\and
Zachary Manchester \\
Robotics Institute \\
Carnegie Mellon University \\
\texttt{zacm@cmu.edu} \\
}

\begin{document}
\maketitle

\begin{abstract}
  \changed{
  We present a data-efficient algorithm for learning models for model-predictive control (MPC). Our approach, Jacobian-Regularized Dynamic-Mode Decomposition (JDMD), offers improved sample efficiency over traditional Koopman approaches based on Dynamic-Mode Decomposition (DMD) by leveraging Jacobian  information from an approximate prior model of the system, and improved tracking performance over traditional model-based MPC. We demonstrate JDMD's ability to quickly learn bilinear Koopman dynamics representations across several realistic examples in simulation, including a perching maneuver for a fixed-wing aircraft with an empirically derived high-fidelity physics model. In all cases, we show that the models learned by JDMD provide superior tracking and generalization performance within a model-predictive control framework, even in the presence of significant model mismatch, when compared to approximate prior models and models learned by standard Extended DMD (EDMD).
  }
\end{abstract}

\section{Introduction}

In recent years, both model-based optimal-control \cite{Farshidian2017,Kuindersma2014,Bjelonic2021,Subosits2019} and data-driven reinforcement-learning methods \cite{Karnchanachari2020,Hoeller2020,Li2021} have demonstrated impressive success on complex, nonlinear robotic systems. However, both approaches suffer from inherent drawbacks: Data-driven methods often require extremely large amounts of data and fail to generalize outside of the domain or task on which they were trained. On the other hand, model-based methods require an accurate model of the system to achieve good performance. In many cases, high-fidelity models can be too difficult to construct from first principles or too computationally expensive to be of practical use. However, low-order approximate models that can be evaluated cheaply at the expense of controller performance are often available. With this in mind, we seek a middle ground between model-based and data-driven approaches in this work. 

We propose a method for learning bilinear Koopman models of nonlinear
dynamical systems for use in model-predictive control that leverages \changed{derivative} information from an
approximate prior dynamics model of the system in the training process.
\changed{Given the increased availability of differentiable simulators \cite{howell2022dojo, todorov2012mujoco},
this approximate derivative information is readily available for many systems of interest.}
Our new algorithm
builds on Extended Dynamic Mode Decomposition (EDMD), which learns Koopman models from
trajectory data \cite{Meduri2022,Bruder2021,Korda2018,Folkestad2020,Suh2020}, by adding a
derivative regularization term based on derivatives computed from a prior model.  We show
that this new algorithm, Jacobian-regularized Dynamic Mode Decomposition (JDMD), can learn
models with dramatically fewer samples than EDMD, even when the prior model differs
significantly from the true dynamics of the system.  We also demonstrate the effectiveness
of these learned models in a model-predictive control (MPC) framework.
The result is a fast, robust, and sample-efficient pipeline for quickly training a model
that can outperform \changed{MPC controllers using both approximate analytical models as well as
models learned using traditional Koopman approaches.}

Our work is most closely related to the recent works of Folkestad et. al.
\cite{Folkestad2020,Folkestad2021,Folkestad2021a}, which learn bilinear models and apply
nonlinear model-predictive control directly on the learned bilinear dynamics. Other recent
works have combined linear Koopman models with model-predictive control \cite{Korda2018} and
Lyapunov control techniques with bilinear Koopman models \cite{Narasingam2022}. Our contributions
are:

\begin{itemize}
  \item A novel extension to Dynamic Mode Decomposition, called JDMD, that
  incorporates gradient information from an approximate analytic model
  
  \item A recursive, batch QR algorithm for solving the least-squares problems that arise 
  when learning bilinear dynamical systems using DMD-based algorithms, including JDMD and EDMD
  
\end{itemize}

The remainder of the paper is organized as follows: In Section
\ref{sec:Preliminaries/Background} we provide some background on the application of Koopman
operator theory to controlled dynamical systems and review some related works.  Section
\ref{sec:jdmd} then describes the proposed JDMD algorithm.  In Section \ref{sec:rls} we
outline a memory-efficient technique for solving the large, sparse linear least-squares
problems that arise when applying JDMD and other DMD-based algorithms.  
Section \ref{sec:results} then
provides simulation results and analysis of the proposed algorithm applied to control tasks
on a cartpole, 
a quadrotor, and a small foam airplane \changed{with an experimentally determined aerodynamics model}, 
all subject to significant model mismatch. 
\changed{We also discuss the suitability of the conventional open-loop prediction error as a metric for evaluating dynamics model used in closed-loop control frameworks.}
In Section \ref{sec:limitations} we discuss the limitations of our approach,
followed by some concluding remarks in Section \ref{sec:conclusion}.

\section{Background and Related Work} \label{sec:Preliminaries/Background}

\subsection{Koopman Operator Theory}

The theoretical underpinnings of the Koopman operator and its application to dynamical
systems has been extensively studied \cite{Bruder2021,brunton2016discovering,BRUNTON2016710,Proctor2018,Williams2015,Surana2016}.
Rather than describe the theory in detail, we highlight the key concepts employed by the
current work and refer the reader to the existing literature on Koopman theory for further
details.

We start by assuming a controlled, nonlinear, discrete-time dynamical system,
\begin{equation} \label{eq:discrete_dynamics} 
  x^+ = f(x, u), 
\end{equation} 
where $x \in \mathcal{X} \subseteq \R^{N_x}$ is the state vector, $u \in \R^{N_u}$ is the
control vector, and $x^+$ is the state at the next time step. \changed{Assuming the dynamics
are control affine, the nonlinear finite-dimensional system \eqref{eq:discrete_dynamics}
can be represented \emph{exactly} by an infinite-dimensional bilinear system through the
Koopman canonical transform \cite{Surana2016}. This bilinear Koopman model takes the form,}
\begin{equation} \label{eq:bilinear_dynamics}
  y^+ = A y + B u + \sum_{i=1}^m u^i C^i y = g(y,u) ,
\end{equation}
where $y = \phi(x)$ is a nonlinear mapping from the finite-dimensional state space
$\mathcal{X}$ to the infinite-dimensional Hilbert space of \textit{observables}
$\mathcal{Y}$.  In practice, we approximate \eqref{eq:bilinear_dynamics} by restricting
$\mathcal{Y}$ to be a finite-dimensional vector space, in which case $\phi$ becomes a
finite-dimensional nonlinear function of the state variables \changed{that can be either chosen 
heuristically based on domain expertise or by learning
\cite{folkestad2020extended,folkestad2022koopnet,li2017extended}}.

Intuitively, $\phi$ ``lifts'' our state $x$ into a higher dimensional space $\mathcal{Y}$
where the dynamics are approximately bilinear, effectively trading dimensionality for
bilinearity. Similarly, we can perform an ``unlifting'' operation by projecting a lifted state
$y$ back into the original state space $\mathcal{X}$. In this work, 
\changed{since we embed the original state within the nonlinear mapping \cite{Bruder2021, Folkestad2021, mamakoukas_local_2019, huang_feedback_2018, ma_optimal_2019}}, 
$\phi$ is constructed in such a way that this unlifting is linear:
\begin{equation}
	x = G y.
\end{equation}
\changed{We note that our proposed method does not require this assumption: any mapping could be used. The 
problem of finding an optimal mapping is itself a major area of research, and many recent studies have focused
on jointly learning both the model and the mapping
\cite{folkestad2020extended,
folkestad2022koopnet,li2017extended,wang2021deep,kaiser2021data}. 
While clearly advantageous, learning
an optimal mapping is not the focus of this paper. Instead, we focus on incorporating prior information from an approximate model in a way that is applicable to any lifting function, and we rely on simple mappings that are chosen heuristically in all of our examples.}


\subsection{Extended Dynamic Mode Decomposition} \label{sec:edmd}

A lifted bilinear system of the form \eqref{eq:bilinear_dynamics} can be learned from $P$
samples of the system dynamics $(x_j^+,x_j,u_j)$ using Extended Dynamic Mode Decomposition
(EDMD) \cite{Folkestad2021,Williams2015}. We first define the following data matrices: 
\begin{equation}
  Z_{1:P} = \begin{bmatrix}
    y_1         & y_2         & \dots  & y_P          \\
    u_1         & u_2         & \dots  & u_P          \\
    u^{1}_{1} y_1 & u^{1}_{2} y_2 & \dots  & u^{1}_{P} y_P  \\
    \vdots      & \vdots      & \ddots & \vdots       \\
    u^{m}_{1} y_1 & u^{m}_{2} y_2 & \dots  & u^{m}_{P} y_P  \\
  \end{bmatrix}, \quad 
  Y_{1:P}^+ = \begin{bmatrix}
    y_1^+         & y_2^+         & \dots  & y_P^+    \\
  \end{bmatrix},
\end{equation}
where $u^{i}_{k}$ is the $i$-th element of the control vector at time $k$.
We then concatenate all of the model coefficient matrices from \eqref{eq:bilinear_dynamics} as follows:
\begin{equation} \label{eq:E_matrixdef}
  E = \begin{bmatrix} A & B & C^1 & \dots & C^m \end{bmatrix} \in \R^{N_y \times N_z}.
\end{equation}
The model learning problem can then be written as the following linear least-squares problem:
\begin{align} \label{opt:edmd}
  \underset{E}{\text{minimize}} \; \norm{E Z_{1:P} - Y_{1:P}^+}_2^2
\end{align}
\changed{EDMD is closely related to classical feature-based machine learning approaches like the ``kernel trick'' used in support vector
machines \cite{kutz2016dynamic}, but extends these ideas to bilinear models of controlled dynamical systems.}

\section{Jacobian-Regularized Dynamic Mode Decomposition} \label{sec:jdmd}

We now present JDMD as a straightforward adaptation of the original EDMD algorithm described
in Section \ref{sec:edmd}. Given $P$ samples of the dynamics $(x_j^+, x_j, u_j)$, and an
approximate discrete-time dynamics model,
\begin{equation}
  x^+ = \tilde{f}(x,u),
\end{equation}
we can evaluate the Jacobians of our approximate model $\tilde{f}$ at each of the sample
points: $\tilde{A}_j = \pdv{\tilde{f}}{x}, \tilde{B}_j = \pdv{\tilde{f}}{u}$. After
choosing a nonlinear mapping $\phi : \R^{N_x} \mapsto \R^{N_y}$ our goal is to find a
bilinear dynamics model \eqref{eq:bilinear_dynamics} that matches the Jacobians of our
approximate model, while also matching our dynamics samples. We accomplish this by 
penalizing differences between the Jacobians of our learned bilinear model with respect to 
the original states $x$ and controls $u$, and the Jacobians we expect from our analytical 
model. These \textit{projected Jacobians} are calculated by differentiating through the 
\textit{projected dynamics}:
\begin{equation} \label{eq:projected_dynamics}
  x^+ = G \left( A \phi(x) + B u + \sum_{i=1}^m u^i C^i \phi(x) \right)  = \bar{f}(x,u).
\end{equation}
Differentiating \eqref{eq:projected_dynamics} with respect to $x$ and $u$ gives us
\begin{subequations} \label{eq:projected_jacobians}
  \begin{align}
    \bar{A}_j &= \pdv{\hat{f}}{x}\left(x_j,u_j\right) 
    = G \left(A + \sum_{i=1}^m u^i_j C^i \right) \Phi(x_j)
    = G E \hat{A}(x_j,u_j) = G E \hat{A}_j ,\\
    \bar{B}_j &= \pdv{\hat{f}}{u}\left(x_j,u_j\right) 
    = G \Big(B + \begin{bmatrix} C^1 x_j & \dots & C^m x_j \end{bmatrix} \Big)
    = G E \hat{B}(x_j,u_j) = G E \hat{B}_j ,
  \end{align}
\end{subequations}
where $\Phi(x) = \pdv*{\phi}{x}$ is the Jacobian of the nonlinear map $\phi$, and
\begin{equation}
  \hat{A}(x,u) =  \begin{bmatrix} 
    I_{N_y} \\ 0 \\ u^1 I_{N_y} \\ u^2 I_{N_y} \\ \vdots \\ u^m I_{N_y} 
  \end{bmatrix} \Phi(x) \in \R^{N_z \times N_x}, \quad
  \hat{B}(x,u) = \begin{bmatrix} 
    0 \\ 
    I_{N_u} \\ 
    [\phi(x) \; 0 \; ... \; 0] \\
    [0 \; \phi(x) \; ... \; 0] \\
    \vdots \\
    [0 \; 0 \; ... \; \phi(x)] \\
  \end{bmatrix} \in \R^{N_z \times N_u}.
\end{equation}

We then solve the following linear least-squares problem:
\begin{align} \label{opt:jdmd}
  \underset{E}{\text{minimize}} \;\; 
    (1-\alpha) \norm{E Z_{1:P} - Y_{1:P}^+}_2^2 + 
        \alpha \sum_{j=1}^P \left( 
          \norm{G E \hat{A}_j - \tilde{A}_j}_2^2 + 
          \norm{G E \hat{B}_j - \tilde{B}_j}_2^2 \right) .
\end{align}

Problem \eqref{opt:jdmd} has $(N_y + N_x^2 + N_x \cdot N_u) \cdot P$ rows
and $N_y \cdot N_z$ columns. Given that the number of rows in this problem grows
quadratically with the state dimension, solving it can be computationally challenging. The next section proposes an algorithm to address this challenge without resorting to a distributed-memory system. This solution method also has the benefit of allowing
incremental updates to the bilinear system, enabling online model learning.

\section{Efficient Recursive Least Squares} \label{sec:rls}
In its canonical formulation, a linear least squares problem can be represented as the
following unconstrained optimization problem:
\begin{align} \label{opt:lls}
  \min_x \|Fx - d\|_2^2.
\end{align}
We assume $F$ is a large, sparse matrix and that solving it directly using a QR or Cholesky
decomposition requires too much memory for a single computer. While solving \eqref{opt:lls}
using an iterative method such as LSMR \cite{Fong2011} or LSQR \cite{Paige1982} is possible,
we find that these methods do not work well in practice for solving \eqref{opt:jdmd} due to
ill-conditioning.  Standard recursive methods for solving these problems are able to process
the rows of the matrices sequentially to build a QR decomposition of the full matrix, but
also tend to suffer from ill-conditioning \cite{Strobach1990,Sayed2009,Ghirnikar1990}.

To overcome these issues, we propose an alternative recursive method based. We solve
\eqref{opt:lls} by dividing rows of $F$ into batches:
\begin{align} \label{eq:F_sum}
  F^T F = F_1^T F_1 + F_2^T F_2 + \ldots + F_N^T F_N.
\end{align}
The main idea is to maintain and update an upper-triangular Cholesky factor $U_i$ of the
first $i$ terms of the sum \eqref{eq:F_sum}. Given $U_i$, we can calculate $U_{i+1}$ using
the $\operatorname{QR}$ decomposition, as shown in
\cite{Howell2019}:
\begin{equation}
  U_{i+1} = \sqrt{U_i^TU_i + F_{i+1}^TF_{i+1}} = 
  \operatorname{QR_R}\bigg( \begin{bmatrix} {U_i} \\ {F_{i+1}} \end{bmatrix} \bigg),
\end{equation}
where $\operatorname{QR_R}$ returns the upper triangular matrix $R$ from the 
$\operatorname{QR}$ decomposition. For an efficient implementation, this function should be
an ``economy'' or ``Q-less'' $\operatorname{QR}$ decomposition 
since the $Q$ matrix is never needed.

We also handle regularization of the normal equations, equivalent to adding quadratic or Tikhonov
regularization to the original least squares problem, during the base case of our recursion,
\begin{equation}
  U_1 =  \operatorname{QR_R}\bigg( 
  \begin{bmatrix} {F_1} \\ \sqrt{\lambda} I \end{bmatrix}
  \bigg),
\end{equation}
where $\lambda$ is a scalar regularization weight.
\changed{To ensure fair comparisons, the results presented in the next section for both EDMD and JDMD correspond to the best-performing $\lambda$ values found by sweeping over a wide parameter range.}

\section{Experimental Results} \label{sec:results}

This section presents the results of several simulation experiments to evaluate the
performance of JDMD. We specify two models for each simulated system: a \textit{nominal}
model, which is simplified and contains both parametric and non-parametric model error, and
a \textit{true} model, which is used exclusively for simulating the system and evaluating
algorithm performance.

All models were trained by simulating the ``true'' system with a nominal controller to 
collect data in the region of the state space relevant to the task. A set of fixed-length 
trajectories were collected, each at a sample rate of 20-25 Hz. The bilinear EDMD model was
trained using the same approach introduced by Folkestad and Burdick \cite{Folkestad2021}.
\changed{When applying MPC to the learned Koopman models, the projected Jacobians 
\eqref{eq:projected_jacobians} were used, since this projected system is much more likely to 
be controllable than the lifted one and reduces the computational complexity of 
the MPC controller. This results in a nonlinear model in the original state space, which 
is linearized about the reference trajectory to create a linear MPC controller.}
All continuous-time dynamics were discretized with an explicit fourth-order Runge-Kutta
method. Code for all experiments is available at \url{https://github.com/bjack205/BilinearControl.jl}.  

\subsection{Systems and Tasks}

\begin{figure}[t!]
    \begin{subfigure}{\linewidth}
        \centering
        \includegraphics[width=\textwidth]{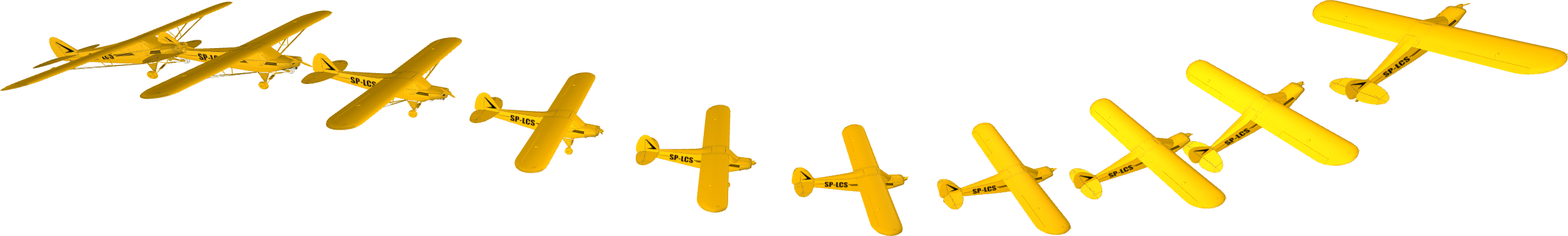}
        \caption{Expert demonstration of a high angle-of-attack perching maneuver that minimizes 
                velocity at the goal position with complex, post-stall aerodynamic forces.}
        \label{fig:perch}
    \end{subfigure}
    \par\medskip
    \begin{subfigure}{.4\linewidth}
        \centering
        \includegraphics[height=2.5cm]{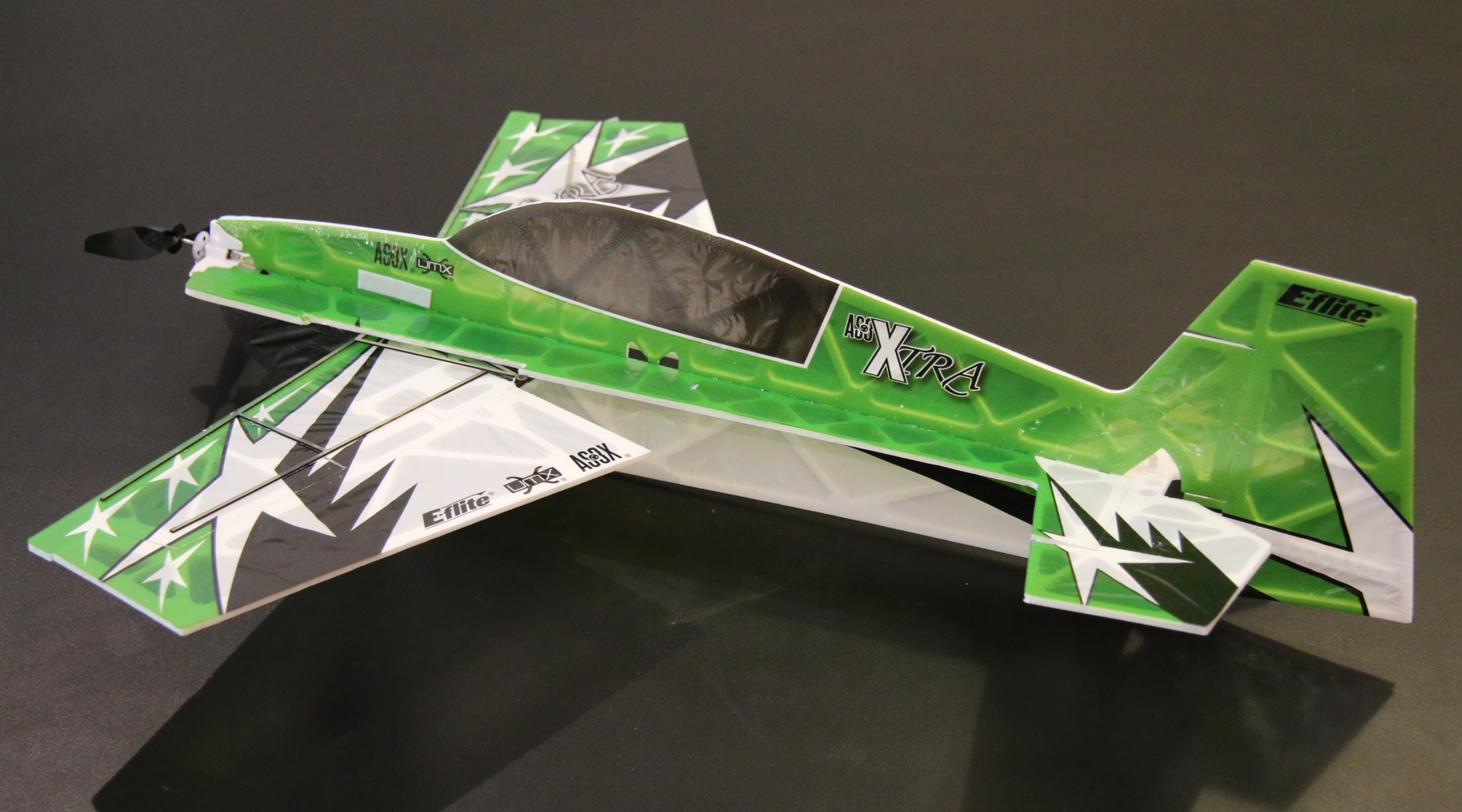}
        \caption{E-Flite AS3Xtra airplane model used in hardware data collection.}
        \label{fig:hw1}
    \end{subfigure}%
    \hfill
    \begin{subfigure}{.52\linewidth}
        \centering
        \includegraphics[height=2.5cm]{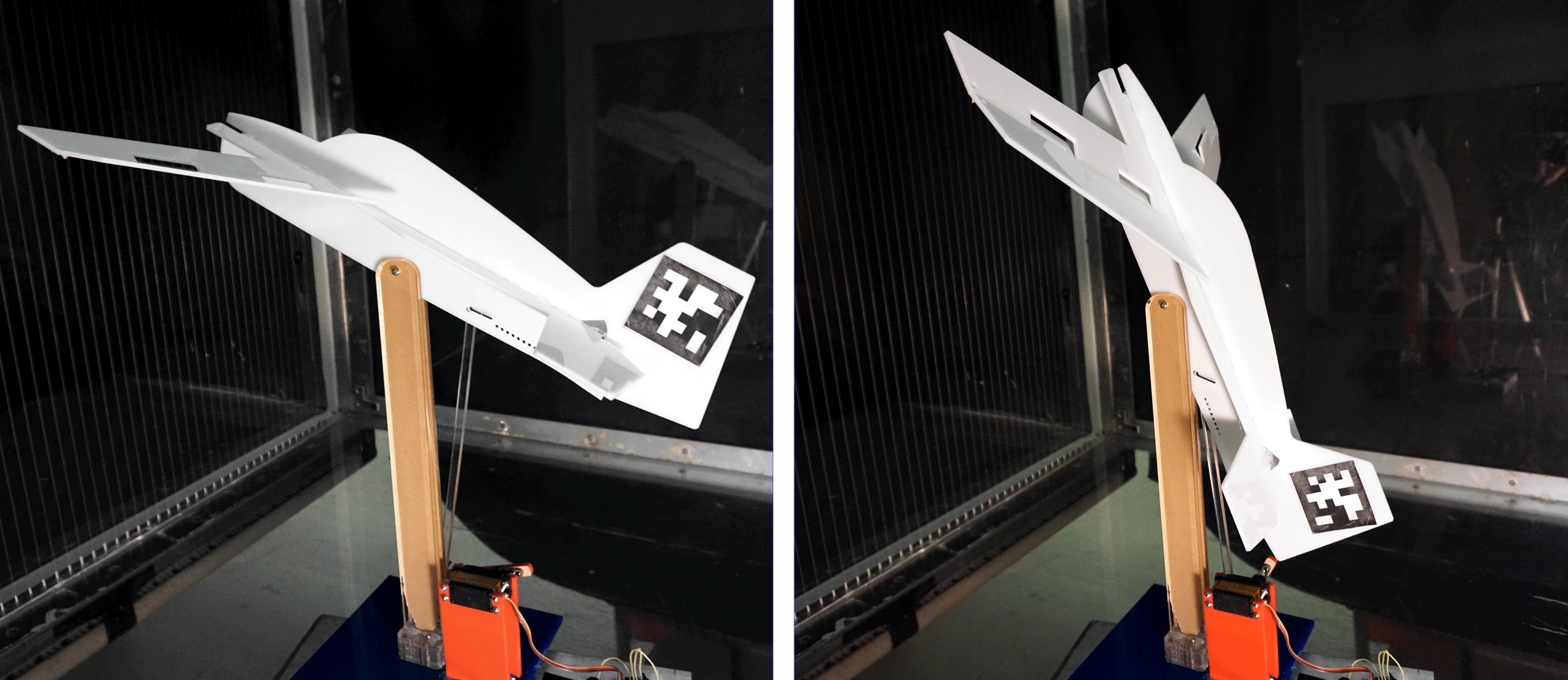}
        \caption{Wind-tunnel experimental setup for collecting aerodynamic data \cite{Manchester2017}.}
        \label{fig:hw2}
    \end{subfigure}\\[1ex]
    \caption{\changed{Complex dynamics of a perching fixed-wing airplane. High-angle-of-attack perching maneuvers (top) require the modeling of complex post-stall aerodynamic effects. The simulated aerodynamic forces were modeled using flight data collected from real-world hardware experiments (bottom) \cite{Manchester2017}.}}
    \label{fig:test}
\end{figure}


\textbf{Cartpole:} We perform a swing-up task on a cartpole system. The \textit{true} model
includes Coulomb friction between the cart and the floor, viscous damping at both joints,
and a deadband in the control input that were not included in the \textit{nominal} model.
Additionally, the mass of the cart and pole model were altered by 20\% and 25\% with respect
to the nominal model, respectively.  The following nonlinear mapping was used when learning
the bilinear models: 
$\phi(x) = [\, 1,\,
x,\, \sin(x),\, \cos(x),\, \sin(2x),\, \sin(4x),\, T_2(x),\, T_3(x),\, T_4(x)\, ] \in
\R^{33}$, where $T_i(x)$ is a Chebyshev polynomial of the first kind of order $i$. 
All reference trajectories for the swing up task were generated using ALTRO 
\cite{Howell2019,Jackson2021}.

\textbf{Quadrotor:} We track point-to-point linear reference trajectories from
various initial conditions on a full 3D quadrotor model. 
The \textit{true} model includes aerodynamic drag terms not included in the \textit{nominal}
model, as well as parametric error of roughly 5\% on the system parameters (e.g. mass, rotor
arm length, etc.). 
The model was trained using a nonlinear mapping of $\phi(x) = [\, 1,\, x,\, T_2(x),\, \sin(p),\,
\cos(p),\, R^{T}v,\ v^{T}RR^{T}v,\, p \times v,\, p \times \omega,\, \omega \times \omega ]
\in \R^{44}$, where $p$ is the quadrotor's position, $v$ and $\omega$ are the translational
and angular velocities respectively, and $R$ is the rotation matrix representation of the quadrotor's attitude.

\textbf{Airplane:} We perform a post-stall perching maneuver on a high-fidelity model of
a fixed-wing airplane. The perching trajectory is produced using trajectory optimization
(see Figure \ref{fig:perch}). \changed{Perching involves flight at high
angles of attack, where the aerodynamic lift and drag forces are extremely complex and
difficult to model from first principles. We leverage previous works that fit post-stall
aerodynamics models using empirical data from wind-tunnel experiments \cite{Manchester2017, Moore2014}.
The \textit{true} model includes these empirical
nonlinear flight dynamics \cite{Manchester2017}, while the \textit{nominal}
model uses a simple flat-plate wing model with linear lift and quadratic drag coefficient
approximations.} The bilinear models use a nonlinear mapping $\phi \in \R^{68}$, which includes
the aircraft attitude (expressed as a vector of Modified Rodriguez Parameters \cite{Schaub09a}),
powers of the angle of attack and side slip angle, the body frame velocity, various cross
products with the angular velocity, and 3rd and 4th order Chebyshev polynomials
of the states.


\subsection{Sensitivity to Model Mismatch}

\begin{wraptable}{R}{4.0cm}
    \vspace{-\baselineskip}
    \centering 
    \begin{tabular}{c|ccc}
        $\mu$ & \rot{60}{1em}{Nominal} & \rot{60}{1em}{EDMD} & \rot{60}{1em}{JDMD} \\
        \midrule
        0.0 & \cmark & 3 & 2 \\ 
        0.1 & \cmark & 19 & 2 \\ 
        0.2 & \xmark & 6 & 2 \\ 
        0.3 & \xmark & 15 & 2 \\ 
        0.4 & \xmark & \xmark & 3 \\ 
        0.5 & \xmark & \xmark & 7 \\ 
        0.6 & \xmark & \xmark & 12 \\ 
    \end{tabular}
  \caption{Training trajectories required to stabilize the cartpole with given friction
    coefficient.
  }
  \vspace{-0.5\baselineskip}
  \label{tab:friction_comp}
\end{wraptable}

While a significant amount of model mismatch is introduced in all examples,
a natural argument against model-based methods is that they are only as good as the model's
ability to capture the salient dynamics of the system. Therefore, we investigated the effect of increasing
model mismatch by incrementally increasing the Coulomb friction coefficient $\mu$ between the cart
and the floor for the cartpole stabilization task (recall the nominal model assumed zero
friction). The results are shown in Table \ref{tab:friction_comp}. As expected, the number
of training trajectories required to find a good stabilizing controller increases with $\mu$. We achieved the results above by setting $\alpha = 0.01$, corresponding 
to a decreased confidence in our model, thereby placing greater weight on the experimental 
data. The standard EDMD approach always required more samples, and was unable to find a good
enough model above friction values of 0.4. While this could likely be remedied by adjusting
the nonlinear mapping $\phi$, the proposed approach works well with the given basis.  Note
that the nominal MPC controller failed to stabilize the system above friction values of 0.1,
so again, we demonstrate that we can improve MPC performance substantially with just a few
training samples by combining analytical derivative information and data sampled from the true
dynamics.


\begin{figure}[b!]
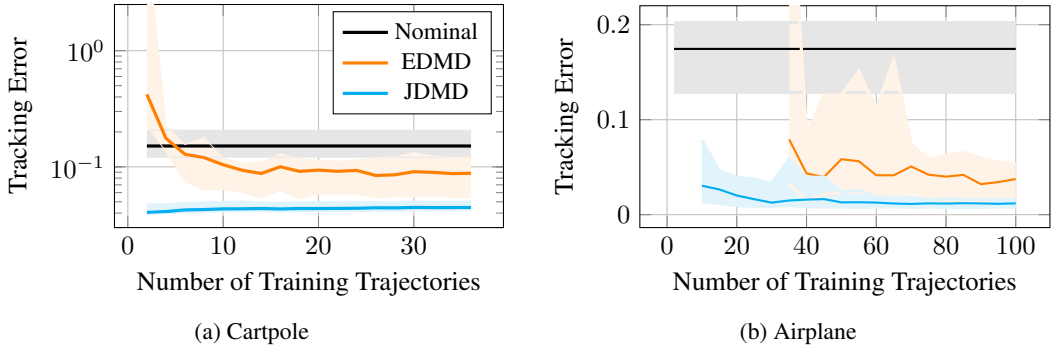

  \centering
  \begin{subfigure}[t]{0.48\textwidth}
    \includegraphics[width=\textwidth, height=4.0cm]{cartpole_mpc_test_error.tikz}
    \caption{Cartpole}
    \label{fig:cartpole_mpc_test_error}
  \end{subfigure}
  \hfill
  \begin{subfigure}[t]{0.48\textwidth}
    \includegraphics[width=\textwidth,height=4.0cm]{airplane_error_by_num_train.tikz}
    \caption{Airplane}
    \label{fig:airplane_sample_efficiency}
  \end{subfigure}
  \caption{MPC tracking error vs training trajectories for both the cartpole (left) and
  airplane (right). Tracking error is defined as the average L2 error over all the
  test trajectories between the reference and simulated trajectories. 
  The median error is shown as a thick line, while the shaded regions represent the 5\% to 95\% bounds on the 10 test trajectories.}
  \label{fig:sample_efficiency}
\end{figure}

\subsection{Sample Efficiency}

\changed{We compare the sample efficiency of several algorithms on the cartpole swing-up task in Figure 
\ref{fig:sample_efficiency}.
As shown, JDMD achieves the best performance overall, and does so
with only two training trajectories. In comparison, traditional EDMD requires about 10 iterations to achieve 
consistent performance.
Similar results were obtained for the airplane perching example (see Figure \ref{fig:airplane_sample_efficiency}),
where EDMD requires over three times the number of samples (35 vs 10) compared to JDMD and never achieves the same closed-loop performance.}
\subsection{Generalization}

\begin{wraptable}{R}{7.0cm}
	\vspace{-2\baselineskip}
	\begin{tabular}{cccc}\\
		\toprule  
		& {\color{black} \textbf{Nominal}} & {\color{orange} \textbf{EDMD}} & {\textbf{\color{cyan} JDMD}} \\
		\midrule
		Success Rate 		& \textbf{82\%} & 18\%	& 80\% \\
		Median      		& 0.30			& 0.63 	& \textbf{0.11} \\
            5\% Quantile        & 0.13          & 0.08 & \textbf{0.03} \\
            95\% Quantile       & 0.38          & 2.62  & \textbf{0.23} \\
		\bottomrule
	\end{tabular}
	\caption{Performance summary of MPC tracking of 6-DOF quadrotor. Other than success rate, all values are the tracking error of the successfully stabilized trajectories.}
	\label{tab:full_quad_tracking_mpc}
\end{wraptable}

\begin{figure}[t] \centering
	\begin{subfigure}[t]{0.49\textwidth}
		\centering
		\includegraphics[width=\textwidth]{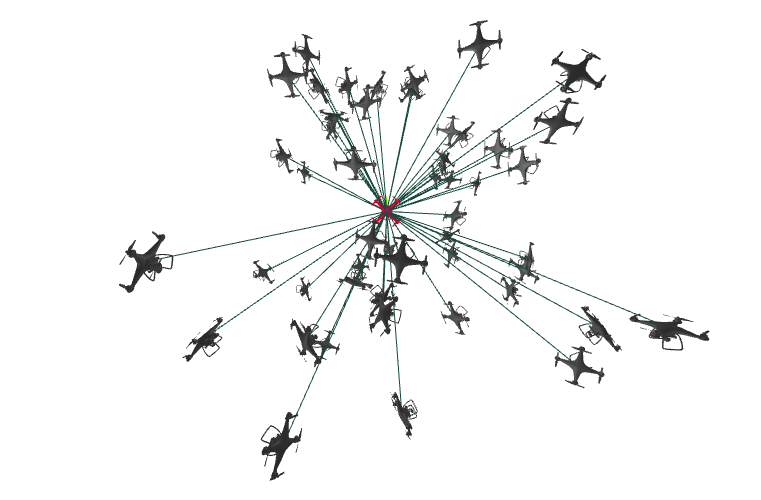}
		\caption{Point-to-point trajectories and initial conditions for testing MPC on a 6-DOF quadrotor.}
		\label{fig:rex_full_quadrotor_initial_conditions}
	\end{subfigure}
	\hfill
	\begin{subfigure}[t]{0.49\textwidth}
		\raggedright
		\includegraphics[width=\textwidth]{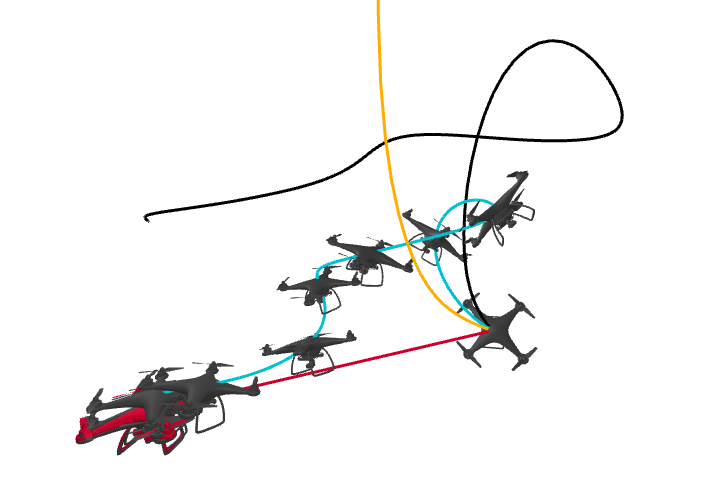}
		\caption{Closed-loop trajectories of nominal MPC (black), EDMD (\color{orange}
		orange\color{black}), and JDMD (\color{cyan} cyan\color{black}) for tracking a dynamically infeasible,
		point-to-point trajectory (\color{red} red\color{black}).}
		\label{fig:jdmd_full_quad_pointtopoint_with_waypoints}
	\end{subfigure}
	\caption{
	Visualizations of the tests on the full quadrotor model.
	}
	\label{fig:training_window}
\end{figure}

We demonstrate the generalizability of JDMD on a 
quadrotor. The task is to return to the origin, given an initial condition 
sampled from a uniform distribution centered at the origin. Test initial conditions are sampled from a distribution larger than that
of the training data.
Given the goal of tracking a straight line back to the origin, we
test 50 initial conditions, many of which are far from the goal, have large velocities, or
are nearly inverted (see Figure \ref{fig:rex_full_quadrotor_initial_conditions}). 
The results using an MPC controller are shown in Table \ref{tab:full_quad_tracking_mpc}, 
demonstrating the generalizability of JDMD, given that the algorithm 
was only trained on 30 initial conditions sampled relatively sparsely given the size of the 
sampling window. EDMD only successfully brings about 18\% of the samples to the origin, 
while the majority of the time resulting in trajectories like those in Figure 
\ref{fig:jdmd_full_quad_pointtopoint_with_waypoints}. \changed{JDMD improves the tracking 
performance of nominal MPC, which is subject to a constant-bias error due to model mismatch, 
as shown in Figure \ref{fig:jdmd_full_quad_pointtopoint_with_waypoints}.}

\begin{figure}
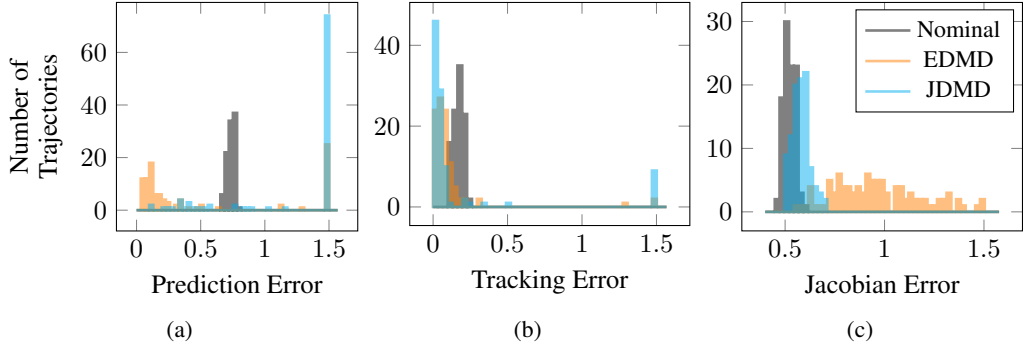

    \centering
    \begin{subfigure}[t]{0.34\linewidth}
        \centering
        \includegraphics[width=\linewidth, height=4.0cm]{airplane_hist_openloop.tikz}
        \caption{}
        \label{fig:airplane_hist_openloop}
    \end{subfigure}
    \begin{subfigure}[t]{0.31\linewidth}
        \centering
        \includegraphics[width=\linewidth, height=4.0cm]{airplane_hist_closedloop.tikz}
        \caption{}
        \label{fig:airplane_hist_closedloop}
    \end{subfigure}
    \begin{subfigure}[t]{0.31\linewidth}
        \centering
        \includegraphics[width=\linewidth, height=4.0cm]{airplane_hist_jac_err.tikz}
        \caption{}
        \label{fig:airplane_hist_jac_err}
    \end{subfigure}
    \caption{Histograms displaying prediction, tracking, and Jacobian error over 100 test trajectories with randomly sampled initial conditions for the airplane perching problem.
    The EDMD model is trained with 24 trajectories and the JDMD model is trained with 8. While JDMD has poor open-loop prediction (a), it has better closed-loop tracking performance (b) and better matching of the dynamics Jacobians (c).} 
\end{figure}

\changed{
\subsection{Model Prediction Error vs. Controller Performance}

Much of the previous literature focuses on open-loop prediction error for evaluating learned-dynamics models \cite{Bruder2021, Korda2018, Folkestad2021, BRUNTON2016710, Surana2016, djeumou2022neural}. While intuitive, we argue that this is a poor metric when the end goal is closed-loop control performance. As shown in the histogram of open-loop prediction error in Figure
\ref{fig:airplane_hist_openloop}, the open-loop prediction error of JDMD (trained with 8 trajectories) is significantly higher over 100 test trajectories, with 74\% of tests resulting in a prediction error of approximately $1.5$ compared to 25\% for EDMD (trained with 24 trajectories). 
Despite worse open-loop prediction performance, the JDMD model outperforms the EDMD model in closed-loop tracking (see Figure \ref{fig:airplane_hist_closedloop}). Given that MPC, like most closed-loop controllers, relies on the behavior of the model under small perturbations (i.e. derivative information), the difference in tracking performance may be explained by JDMD achieving a distribution with much lower Jacobian error than EDMD (see Figure \ref{fig:airplane_hist_jac_err}). This suggests that open-loop prediction error is not necessarily a good metric for evaluating models that will be used in control applications, and that models sufficient for closed-loop control may be learned with far less data.

}

\section{Limitations} \label{sec:limitations}

\changed{Many of the limitations of the proposed approach derive from the limitations of
Koopman approaches more broadly, such as the sensitivity to the nonlinear mapping selected
and the limitation to control-affine continuous dynamics. 
While the presented single rigid-body
systems such as the quadrotor or airplane have similar dimensionality to many autonomous
systems of interest, extensions to systems with many degrees of freedom may be difficult
computationally, given that Jacobian matrices grow in size with the square of the state dimension.
} 
As with most data-driven techniques, it is difficult to claim that our
method will increase performance in all cases. It is possible that having an extremely poor
prior model may hurt rather than help the training process, especially if the derivative 
information from the approximate model has the incorrect sign.

\section{Conclusions and Future Work} \label{sec:conclusion}

We have presented JDMD, a simple but powerful extension to EDMD that incorporates derivative
information from an approximate prior model. We have tested JDMD in combination with 
a simple linear MPC control policy across a range of systems and tasks, and have found that the
resulting combination can dramatically increase sample efficiency over EDMD, often improving over a
nominal MPC policy with just a few sample trajectories. \changed{We also argued that the conventional open-loop
dynamics prediction error is a poor metric for evaluating models used in closed-loop control frameworks.}
Substantial areas for future work
remain: most notably, demonstrating the proposed pipeline on hardware. Additional directions
include \changed{applications on sytems with many degrees of freedom such as those whose dynamics are 
governed by discretized PDEs}, online learning or adaptive control applications, combining simulated and real
data through the use of modern differentiable physics engines \cite{todorov2012mujoco, howell2022dojo},
residual dynamics learning, as well as the development of specialized numerical methods for
solving nonlinear optimal control problems using the learned bilinear dynamics.

\newpage
\bibliography{Koopman.bib}

\begin{thebibliography}{42}
\providecommand{\natexlab}[1]{#1}
\providecommand{\url}[1]{\texttt{#1}}
\expandafter\ifx\csname urlstyle\endcsname\relax
  \providecommand{\doi}[1]{doi: #1}\else
  \providecommand{\doi}{doi: \begingroup \urlstyle{rm}\Url}\fi

\bibitem[Farshidian et~al.()Farshidian, Neunert, Winkler, Rey, and
  Buchli]{Farshidian2017}
F.~Farshidian, M.~Neunert, A.~W. Winkler, G.~Rey, and J.~Buchli.
\newblock An efficient optimal planning and control framework for quadrupedal
  locomotion.
\newblock In \emph{2017 \{\vphantom\}{{IEEE}}\vphantom\{\} {{International
  Conference}} on {{Robotics}} and {{Automation}}
  (\{\vphantom\}{{ICRA}}\vphantom\{\})}, pages 93--100.
\newblock \doi{10.1109/ICRA.2017.7989016}.

\bibitem[Kuindersma et~al.()Kuindersma, Permenter, and Tedrake]{Kuindersma2014}
S.~Kuindersma, F.~Permenter, and R.~Tedrake.
\newblock An efficiently solvable quadratic program for stabilizing dynamic
  locomotion.
\newblock pages 2589--2594.
\newblock ISSN 9781479936854.
\newblock \doi{10.1109/ICRA.2014.6907230}.

\bibitem[Bjelonic et~al.()Bjelonic, Grandia, Harley, Galliard, Zimmermann, and
  Hutter]{Bjelonic2021}
M.~Bjelonic, R.~Grandia, O.~Harley, C.~Galliard, S.~Zimmermann, and M.~Hutter.
\newblock Whole-{{Body MPC}} and {{Online Gait Sequence Generation}} for
  {{Wheeled-Legged Robots}}.
\newblock pages 8388--8395.
\newblock ISSN 9781665417143.
\newblock \doi{10.1109/IROS51168.2021.9636371}.

\bibitem[Subosits and Gerdes()]{Subosits2019}
J.~K. Subosits and J.~C. Gerdes.
\newblock From the racetrack to the road: {{Real-time}} trajectory replanning
  for autonomous driving.
\newblock 4\penalty0 (2):\penalty0 309--320.
\newblock \doi{10.1109/TIV.2019.2904390}.

\bibitem[Karnchanachari et~al.()Karnchanachari, Valls, David~Hoeller, and
  Hutter]{Karnchanachari2020}
N.~Karnchanachari, M.~I. Valls, S.~David~Hoeller, and M.~Hutter.
\newblock Practical {{Reinforcement Learning For MPC}}: {{Learning}} from
  sparse objectives in under an hour on a real robot.
\newblock pages 1--14.
\newblock \doi{10.3929/ETHZ-B-000404690}.
\newblock URL \url{https://doi.org/10.3929/ethz-b-000404690}.

\bibitem[Hoeller et~al.()Hoeller, Farshidian, Hutter, Farshidian, and
  Hoeller]{Hoeller2020}
D.~. Hoeller, F.~. Farshidian, M.~Hutter, F.~Farshidian, and D.~Hoeller.
\newblock Deep {{Value Model Predictive Control}}.
\newblock 100:\penalty0 990--1004.
\newblock \doi{10.3929/ETHZ-B-000368961}.
\newblock URL \url{https://doi.org/10.3929/ethz-b-000368961}.

\bibitem[Li et~al.()Li, Cheng, Peng, Abbeel, Levine, Berseth, and
  Sreenath]{Li2021}
Z.~Li, X.~Cheng, X.~B. Peng, P.~Abbeel, S.~Levine, G.~Berseth, and K.~Sreenath.
\newblock Reinforcement {{Learning}} for {{Robust Parameterized Locomotion
  Control}} of {{Bipedal Robots}}.
\newblock 2021-May:\penalty0 2811--2817.
\newblock ISSN 9781728190778.
\newblock \doi{10.1109/ICRA48506.2021.9560769}.

\bibitem[Howell et~al.(2022)Howell, Cleac'h, Kolter, Schwager, and
  Manchester]{howell2022dojo}
T.~A. Howell, S.~L. Cleac'h, J.~Z. Kolter, M.~Schwager, and Z.~Manchester.
\newblock Dojo: A differentiable simulator for robotics.
\newblock \emph{arXiv preprint arXiv:2203.00806}, 2022.

\bibitem[Todorov et~al.(2012)Todorov, Erez, and Tassa]{todorov2012mujoco}
E.~Todorov, T.~Erez, and Y.~Tassa.
\newblock Mujoco: A physics engine for model-based control.
\newblock In \emph{2012 IEEE/RSJ International Conference on Intelligent Robots
  and Systems}, pages 5026--5033. IEEE, 2012.
\newblock \doi{10.1109/IROS.2012.6386109}.

\bibitem[Meduri et~al.()Meduri, Shah, Viereck, Khadiv, Havoutis, and
  Righetti]{Meduri2022}
A.~Meduri, P.~Shah, J.~Viereck, M.~Khadiv, I.~Havoutis, and L.~Righetti.
\newblock {{BiConMP}}: {{A Nonlinear Model Predictive Control Framework}} for
  {{Whole Body Motion Planning}}.
\newblock \doi{10.48550/arxiv.2201.07601}.
\newblock URL \url{https://arxiv.org/abs/2201.07601v1}.

\bibitem[Bruder et~al.()Bruder, Fu, and Vasudevan]{Bruder2021}
D.~Bruder, X.~Fu, and R.~Vasudevan.
\newblock Advantages of {{Bilinear Koopman Realizations}} for the {{Modeling}}
  and {{Control}} of {{Systems}} with {{Unknown Dynamics}}.
\newblock 6\penalty0 (3):\penalty0 4369--4376.
\newblock \doi{10.1109/LRA.2021.3068117}.

\bibitem[Korda and Mezić()]{Korda2018}
M.~Korda and I.~Mezić.
\newblock Linear predictors for nonlinear dynamical systems: {{Koopman}}
  operator meets model predictive control.
\newblock 93:\penalty0 149--160.
\newblock \doi{10.1016/j.automatica.2018.03.046}.
\newblock URL \url{https://doi.org/10.1016/j.automatica.2018.03.046}.

\bibitem[Folkestad et~al.()Folkestad, Pastor, and Burdick]{Folkestad2020}
C.~Folkestad, D.~Pastor, and J.~W. Burdick.
\newblock Episodic {{Koopman Learning}} of {{Nonlinear Robot Dynamics}} with
  {{Application}} to {{Fast Multirotor Landing}}.
\newblock pages 9216--9222.
\newblock ISSN 9781728173955.
\newblock \doi{10.1109/ICRA40945.2020.9197510}.

\bibitem[Suh and Tedrake()]{Suh2020}
H.~J. Suh and R.~Tedrake.
\newblock The {{Surprising Effectiveness}} of {{Linear Models}} for {{Visual
  Foresight}} in {{Object Pile Manipulation}}.
\newblock 17:\penalty0 347--363.
\newblock \doi{10.48550/arxiv.2002.09093}.
\newblock URL \url{https://arxiv.org/abs/2002.09093v3}.

\bibitem[Folkestad and Burdick()]{Folkestad2021}
C.~Folkestad and J.~W. Burdick.
\newblock Koopman {{NMPC}}: {{Koopman-based Learning}} and {{Nonlinear Model
  Predictive Control}} of {{Control-affine Systems}}.
\newblock In \emph{Proceedings - {{IEEE International Conference}} on
  {{Robotics}} and {{Automation}}}, volume 2021-May, pages 7350--7356.
  {Institute of Electrical and Electronics Engineers Inc.}
\newblock ISBN 978-1-72819-077-8.
\newblock \doi{10.1109/ICRA48506.2021.9562002}.

\bibitem[Folkestad et~al.()Folkestad, Wei, and Burdick]{Folkestad2021a}
C.~Folkestad, S.~X. Wei, and J.~W. Burdick.
\newblock Quadrotor {{Trajectory Tracking}} with {{Learned Dynamics}}: {{Joint
  Koopman-based Learning}} of {{System Models}} and {{Function Dictionaries}}.
\newblock URL \url{http://arxiv.org/abs/2110.10341}.

\bibitem[Narasingam et~al.()Narasingam, Sang, and Kwon]{Narasingam2022}
A.~Narasingam, J.~Sang, and I.~Kwon.
\newblock Data-driven feedback stabilization of nonlinear systems:
  {{Koopman-based}} model predictive control.
\newblock pages 1--12.

\bibitem[Brunton et~al.(2016{\natexlab{a}})Brunton, Proctor, and
  Kutz]{brunton2016discovering}
S.~L. Brunton, J.~L. Proctor, and J.~N. Kutz.
\newblock Discovering governing equations from data by sparse identification of
  nonlinear dynamical systems.
\newblock \emph{Proceedings of the national academy of sciences}, 113\penalty0
  (15):\penalty0 3932--3937, 2016{\natexlab{a}}.

\bibitem[Brunton et~al.(2016{\natexlab{b}})Brunton, Proctor, and
  Kutz]{BRUNTON2016710}
S.~L. Brunton, J.~L. Proctor, and J.~N. Kutz.
\newblock Sparse identification of nonlinear dynamics with control (sindyc).
\newblock \emph{IFAC-PapersOnLine}, 49\penalty0 (18):\penalty0 710--715,
  2016{\natexlab{b}}.
\newblock ISSN 2405-8963.
\newblock \doi{https://doi.org/10.1016/j.ifacol.2016.10.249}.
\newblock URL
  \url{https://www.sciencedirect.com/science/article/pii/S2405896316318298}.
\newblock 10th IFAC Symposium on Nonlinear Control Systems NOLCOS 2016.

\bibitem[Proctor et~al.()Proctor, Brunton, and Nathan~Kutz]{Proctor2018}
J.~L. Proctor, S.~L. Brunton, and J.~Nathan~Kutz.
\newblock Generalizing koopman theory to allow for inputs and control.
\newblock 17\penalty0 (1):\penalty0 909--930.
\newblock \doi{10.1137/16M1062296}.
\newblock URL \url{http://www.siam.org/journals/siads/17-1/M106229.html}.

\bibitem[Williams et~al.()Williams, Kevrekidis, and Rowley]{Williams2015}
M.~O. Williams, I.~G. Kevrekidis, and C.~W. Rowley.
\newblock A {{Data}}–{{Driven Approximation}} of the {{Koopman Operator}}:
  {{Extending Dynamic Mode Decomposition}}.
\newblock 25\penalty0 (6):\penalty0 1307--1346.
\newblock \doi{10.1007/S00332-015-9258-5/FIGURES/14}.
\newblock URL
  \url{https://link.springer.com/article/10.1007/s00332-015-9258-5}.

\bibitem[Surana()]{Surana2016}
A.~Surana.
\newblock \emph{Koopman Operator Based Observer Synthesis for Control-Affine
  Nonlinear Systems; {{Koopman}} Operator Based Observer Synthesis for
  Control-Affine Nonlinear Systems}.
\newblock ISBN 978-1-5090-1837-6.
\newblock \doi{10.1109/CDC.2016.7799268}.

\bibitem[Folkestad et~al.(2020)Folkestad, Pastor, Mezic, Mohr, Fonoberova, and
  Burdick]{folkestad2020extended}
C.~Folkestad, D.~Pastor, I.~Mezic, R.~Mohr, M.~Fonoberova, and J.~Burdick.
\newblock {E}xtended {D}ynamic {M}ode {D}ecomposition with {L}earned {K}oopman
  {E}igenfunctions for {P}rediction and {C}ontrol.
\newblock In \emph{2020 American Control Conference (ACC)}, pages 3906--3913.
  IEEE, 2020.

\bibitem[Folkestad et~al.(2022)Folkestad, Wei, and
  Burdick]{folkestad2022koopnet}
C.~Folkestad, S.~X. Wei, and J.~W. Burdick.
\newblock Koopnet: Joint learning of koopman bilinear models and function
  dictionaries with application to quadrotor trajectory tracking.
\newblock In \emph{2022 International Conference on Robotics and Automation
  (ICRA)}, pages 1344--1350. IEEE, 2022.

\bibitem[Li et~al.(2017)Li, Dietrich, Bollt, and Kevrekidis]{li2017extended}
Q.~Li, F.~Dietrich, E.~M. Bollt, and I.~G. Kevrekidis.
\newblock Extended dynamic mode decomposition with dictionary learning: A
  data-driven adaptive spectral decomposition of the koopman operator.
\newblock \emph{Chaos: An Interdisciplinary Journal of Nonlinear Science},
  27\penalty0 (10):\penalty0 103111, 2017.

\bibitem[Mamakoukas et~al.(2019)Mamakoukas, Castano, Tan, and
  Murphey]{mamakoukas_local_2019}
G.~Mamakoukas, M.~Castano, X.~Tan, and T.~Murphey.
\newblock Local {Koopman} operators for data-driven control of robotic systems.
\newblock In \emph{Robotics: science and systems}, 2019.

\bibitem[Huang et~al.(2018)Huang, Ma, and Vaidya]{huang_feedback_2018}
B.~Huang, X.~Ma, and U.~Vaidya.
\newblock Feedback stabilization using koopman operator.
\newblock In \emph{2018 {IEEE} {Conference} on {Decision} and {Control}
  ({CDC})}, pages 6434--6439. IEEE, 2018.

\bibitem[Ma et~al.(2019)Ma, Huang, and Vaidya]{ma_optimal_2019}
X.~Ma, B.~Huang, and U.~Vaidya.
\newblock Optimal quadratic regulation of nonlinear system using koopman
  operator.
\newblock In \emph{2019 {American} {Control} {Conference} ({ACC})}, pages
  4911--4916. IEEE, 2019.

\bibitem[Wang et~al.(2021)Wang, Han, and Vaidya]{wang2021deep}
R.~Wang, Y.~Han, and U.~Vaidya.
\newblock Deep koopman data-driven optimal control framework for autonomous
  racing.
\newblock \emph{Early Access}, 5, 2021.

\bibitem[Kaiser et~al.(2021)Kaiser, Kutz, and Brunton]{kaiser2021data}
E.~Kaiser, J.~N. Kutz, and S.~L. Brunton.
\newblock Data-driven discovery of koopman eigenfunctions for control.
\newblock \emph{Machine Learning: Science and Technology}, 2\penalty0
  (3):\penalty0 035023, 2021.

\bibitem[Kutz et~al.(2016)Kutz, Brunton, Brunton, and Proctor]{kutz2016dynamic}
J.~N. Kutz, S.~L. Brunton, B.~W. Brunton, and J.~L. Proctor.
\newblock \emph{Dynamic mode decomposition: data-driven modeling of complex
  systems}.
\newblock SIAM, 2016.

\bibitem[Fong and Saunders()]{Fong2011}
D.~C.-L. Fong and M.~Saunders.
\newblock {{LSMR}}: {{An Iterative Algorithm}} for {{Sparse Least-Squares
  Problems}}.
\newblock 33\penalty0 (5):\penalty0 2950--2971.
\newblock ISSN 1064-8275.
\newblock \doi{10.1137/10079687X}.
\newblock URL \url{https://epubs.siam.org/doi/abs/10.1137/10079687X}.

\bibitem[Paige and Saunders()]{Paige1982}
C.~C. Paige and M.~A. Saunders.
\newblock {{LSQR}}: {{An Algorithm}} for {{Sparse Linear Equations}} and
  {{Sparse Least Squares}}.
\newblock 8\penalty0 (1):\penalty0 43--71.
\newblock ISSN 0098-3500, 1557-7295.
\newblock \doi{10.1145/355984.355989}.
\newblock URL \url{https://dl.acm.org/doi/10.1145/355984.355989}.

\bibitem[Strobach()]{Strobach1990}
P.~Strobach.
\newblock Recursive {{Least-Squares Using}} the {{QR Decomposition}}.
\newblock In P.~Strobach, editor, \emph{Linear {{Prediction Theory}}: {{A
  Mathematical Basis}} for {{Adaptive Systems}}}, Springer {{Series}} in
  {{Information Sciences}}, pages 63--101. {Springer}.
\newblock ISBN 978-3-642-75206-3.
\newblock \doi{10.1007/978-3-642-75206-3_4}.
\newblock URL \url{https://doi.org/10.1007/978-3-642-75206-3_4}.

\bibitem[Sayed and Kailath()]{Sayed2009}
A.~Sayed and T.~Kailath.
\newblock \emph{Recursive {{Least-Squares Adaptive Filters}}}, volume 20094251
  of \emph{Electrical {{Engineering Handbook}}}, pages 1--40.
\newblock {CRC Press}.
\newblock ISBN 978-1-4200-4606-9 978-1-4200-4607-6.
\newblock \doi{10.1201/9781420046076-c21}.
\newblock URL
  \url{http://www.crcnetbase.com/doi/abs/10.1201/9781420046076-c21}.

\bibitem[Ghirnikar and Alexander()]{Ghirnikar1990}
A.~Ghirnikar and S.~Alexander.
\newblock Stable recursive least squares filtering using an inverse {{QR}}
  decomposition.
\newblock In \emph{International {{Conference}} on {{Acoustics}}, {{Speech}},
  and {{Signal Processing}}}, pages 1623--1626 vol.3.
\newblock \doi{10.1109/ICASSP.1990.115736}.

\bibitem[Howell et~al.()Howell, Jackson, and Manchester]{Howell2019}
T.~A. Howell, B.~E. Jackson, and Z.~Manchester.
\newblock {{ALTRO}}: {{A Fast Solver}} for {{Constrained Trajectory
  Optimization}}.
\newblock pages 7674--7679.
\newblock ISSN 9781728140049.
\newblock \doi{10.1109/IROS40897.2019.8967788}.

\bibitem[Manchester et~al.()Manchester, Lipton, Wood, and
  Kuindersma]{Manchester2017}
Z.~Manchester, J.~Lipton, R.~Wood, and S.~Kuindersma.
\newblock A {{Variable Forward-Sweep Wing Design}} for {{Enhanced Perching}} in
  {{Micro Aerial Vehicles}}.
\newblock In \emph{{{AIAA Aerospace Sciences Meeting}}}.
\newblock URL \url{https://rexlab.stanford.edu/papers/Morphing_Wing.pdf}.

\bibitem[Jackson et~al.()Jackson, Punnoose, Neamati, Tracy, Jitosho, and
  Manchester]{Jackson2021}
B.~E. Jackson, T.~Punnoose, D.~Neamati, K.~Tracy, R.~Jitosho, and
  Z.~Manchester.
\newblock {{ALTRO-C}}: {{A Fast Solver}} for {{Conic Model-Predictive
  Control}}; {{ALTRO-C}}: {{A Fast Solver}} for {{Conic Model-Predictive
  Control}}.
\newblock ISSN 9781728190778.
\newblock \doi{10.1109/ICRA48506.2021.9561438}.
\newblock URL \url{https://github.com/}.

\bibitem[Moore et~al.(2014)Moore, Cory, and Tedrake]{Moore2014}
J.~Moore, R.~Cory, and R.~Tedrake.
\newblock Robust post-stall perching with a simple fixed-wing glider using
  {LQR}-{Trees}.
\newblock \emph{Bioinspiration \& Biomimetics}, 9\penalty0 (2):\penalty0
  025013, May 2014.
\newblock \doi{10.1088/1748-3182/9/2/025013}.
\newblock URL \url{https://doi.org/10.1088/1748-3182/9/2/025013}.
\newblock Publisher: IOP Publishing.

\bibitem[Schaub and Junkins(2009)]{Schaub09a}
H.~Schaub and J.~Junkins.
\newblock \emph{Analytical {{Mechanics}} of {{Space Systems}}}.
\newblock {{AIAA Education Series}}. {AIAA}, {Reston, VA}, second edition,
  2009.
\newblock ISBN 1-60086-721-9.

\bibitem[Djeumou et~al.(2022)Djeumou, Neary, Goubault, Putot, and
  Topcu]{djeumou2022neural}
F.~Djeumou, C.~Neary, E.~Goubault, S.~Putot, and U.~Topcu.
\newblock Neural networks with physics-informed architectures and constraints
  for dynamical systems modeling.
\newblock In \emph{Learning for Dynamics and Control Conference}, pages
  263--277. PMLR, 2022.

\end{thebibliography}

\end{document}